\documentclass{article} % For LaTeX2e
\usepackage{nips15submit_e,times}
\usepackage{hyperref}
\usepackage{url}
\usepackage{natbib}
\usepackage{amsmath}
\usepackage{amsthm}
\usepackage{amssymb}
\usepackage{graphicx}
\usepackage{xspace}
\usepackage{tabularx}
\usepackage{multicol}
\usepackage{multirow}
\usepackage{url}
\usepackage{natbib}
\usepackage{wrapfig}
\usepackage{comment}
\usepackage{listings}
\usepackage{color}
\usepackage[utf8]{inputenc}
\usepackage{fancyvrb}
\usepackage{enumitem}
\usepackage{booktabs}

\usepackage[normalem]{ulem}

\newcommand{\bmx}[0]{\begin{bmatrix}}
\newcommand{\emx}[0]{\end{bmatrix}}

\newcommand{\RR}[0]{\mathbb{R}}

\graphicspath{ {./figures/} }

\title{ReNet: A Recurrent Neural Network Based Alternative to Convolutional Networks}

\author{
    Francesco Visin$^{\star}$ \\
    Politecnico di Milano
\And
    Kyle Kastner$^{\star}$ \\
    University of Montreal
\And
    Kyunghyun Cho$^{\star}$ \\
    University of Montreal
\And
    Matteo Matteucci \\
    Politecnico di Milano \\
\And
    Aaron Courville \\
    University of Montreal \\
\And
    Yoshua Bengio \\
    University of Montreal \\
    CIFAR Senior Fellow
}

\nipsfinalcopy % Uncomment for camera-ready version

\begin{document}

{
\let\thefootnote\relax\footnote{$\star$ Equal contribution}
}

\maketitle

\begin{abstract}
    In this paper, we propose a deep neural network architecture for object
    recognition based on recurrent neural networks. The proposed network, called
    ReNet, replaces the ubiquitous convolution+pooling layer of the deep
    convolutional neural network with four recurrent neural networks that sweep
    horizontally and vertically in both directions across the image. We evaluate
    the proposed ReNet on three widely-used benchmark datasets; MNIST, CIFAR-10 
    and SVHN. The result suggests that ReNet is a viable alternative to the deep
    convolutional neural network, and that further investigation is needed.
\end{abstract}

\section{Introduction}
Convolutional neural networks~\cite[CNN,][]{Fukushima80,LeCun89} have become the
method of choice for object recognition~\citep[see, e.g.,][]{Krizhevsky-2012}.
They have proved to be successful at a variety of benchmark problems including,
but not limited to, handwritten digit recognition~\citep[see,
e.g.,][]{Ciresan-2012}, natural image classification~\citep[see,
e.g.,][]{Lin2014,Simonyan2015,szegedy2014going}, house number 
recognition~\citep[see, e.g.,][]{Goodfellow+et+al-ICLR2014a}, traffic sign
recognition~\citep[see, e.g.,][]{Ciresan-et-al-2012}, as well as for speech 
recognition~\citep[see, e.g.,][]{Hamid2012, sainath2013, toth2014combining}. 
Furthermore, image representations from CNNs trained to recognize objects on a 
large set of more than one million images~\citep{Simonyan2015,szegedy2014going} 
have been found to be extremely helpful in performing other computer vision
tasks such as image caption generation~\citep[see,
e.g.,][]{Vinyals-et-al-arxiv2014,Xu-et-al-arxiv2015}, video description
generation~\citep[see, e.g.,][]{Li2015} and object
localization/detection~\citep[see, e.g.,][]{Sermanet14}.

While the CNN has been especially successful in computer vision, recurrent
neural networks (RNN) have become the method of choice for modeling sequential
data, such as text and sound. Natural language processing (NLP) applications
include language modeling~\citep[see, e.g.,][]{Mikolov-thesis-2012}, and machine
translation~\citep{Sutskever-et-al-NIPS2014,Cho2014,bahdanau2014neural}.  Other
popular areas of application include offline handwriting
recognition/generation~\citep{Graves+Schmidhuber-2009,Graves-et-al-NIPS2007,Graves-arxiv2013}
and speech recognition~\citep{Chorowski-et-al-arxiv2014,Graves+Jaitly-ICML2014}.
RNNs have also been used together with CNNs in speech
recognition~\citep{sainath2015}.  The recent revival of RNNs has largely been
due to advances in learning
algorithms~\citep{Pascanu+al-ICML2013-small,Martens+Sutskever-ICML2011} and
model
architectures~\citep{Pascanu-et-al-ICLR2014,Hochreiter+Schmidhuber-1997,Cho2014}.

%RNNs have previously been extended to multi-dimensional tasks and applied to
%object recognition~\citep{Graves+Schmidhuber-2009} and texture
%classification~\citep{byeon2014texture}. 

The architecture proposed here is
related and inspired by this earlier work, but our model relies on purely
uni-dimensional RNNs coupled in a novel way, rather than on a multi-dimensional
RNN. The basic idea behind the proposed ReNet architecture is to replace each
convolutional layer (with convolution+pooling making up a layer) in the CNN with
four RNNs that sweep over lower-layer features in different directions: (1)
bottom to top, (2) top to bottom, (3) left to right and (4) right to left.
% This recurrent layer ensures that each feature activation from the
% layer is an activation at the specific location with respect to the whole image,
% in contrast to the usual convolution+pooling layer which only has the local
% context window. 
The recurrent layer ensures that each feature activation in its output is an
activation at the specific location \emph{with respect to the whole image}, in
contrast to the usual convolution+pooling layer which only has a local context
window.  The lowest layer of the model sweeps over the input image, with
subsequent layers operating on extracted representations from the layer below,
forming a hierarchical representation of the input.

\citet{Graves+Schmidhuber-2009} have demonstrated an RNN-based object
recognition system for offline Arabic handwriting recognition.  The main
difference between ReNet and the model of \citet{Graves+Schmidhuber-2009} is
that we use the usual sequence RNN, instead of the multidimensional RNN. We make
the latter two parts of a single layer, usually (horizontal) RNNs or one
(horizontal) bidirectional RNN, work on the hidden states computed by the first
two (vertical) RNNs, or one (vertical) bidirectional RNN. This allows us to use
a plain RNN, instead of the more complex multidimensional RNN, while making each
output activation of the layer be computed with respect to the whole input
image.  

One important consequence of the proposed approach compared to the
multidimensional RNN is that the number of RNNs at each layer scales now linearly
with respect to the number of dimensions $d$ of the input image ($2d$). A
multidimensional RNN, on the other hand, requires the exponential number of RNNs
at each layer ($2^d$). Furthermore, the proposed variant is more easily
parallelizable, as each RNN is dependent only along a horizontal or vertical
sequence of patches. This architectural distinction results in our model being
much more amenable to distributed computing than that of
\citet{Graves+Schmidhuber-2009}. 

In this work, we test the proposed ReNet on several widely used object
recognition benchmarks, namely  MNIST~\citep{Lecun99objectrecognition},
CIFAR-10~\citep{KrizhevskyHinton2009} and SVHN~\citep{Netzer-wkshp-2011}. Our
experiments reveal that the model performs comparably to convolutional neural
networks on all these datasets,
%with only cursory expoloration of hyperparameters
suggesting the potential of RNNs as a competitive alternative to CNNs for image
related tasks.

\section{Model Description}

\begin{wrapfigure}{R}{0.3\textwidth}
    \centering
    \includegraphics[width=0.3\columnwidth]{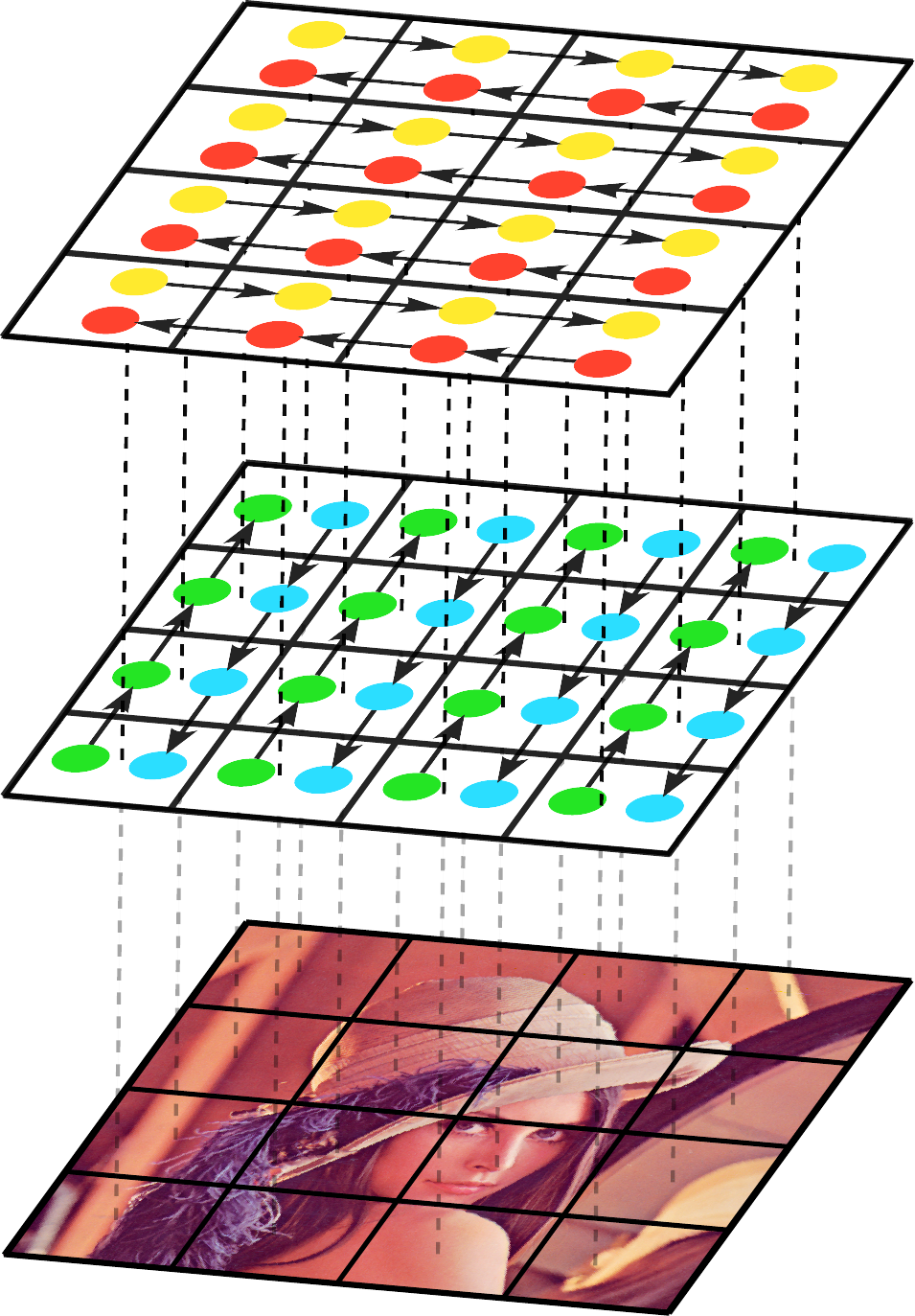}
    \caption{A one-layer ReNet}
    \label{fig:networklayer}
    \vspace{-3mm}
\end{wrapfigure}

Let us denote by $X=\left\{x_{i,j}\right\}$ the input image or the feature map 
from the layer below, where $X \in \RR^{w \times h \times c}$ with $w$, $h$ and 
$c$ the width, height and number of channels, or the feature dimensionality, 
respectively. Given a receptive field (or patch) size of $w_p \times h_p$, we 
split the input image $X$ into a set of $I \times J$ (non-overlapping) patches 
$P = \left\{ p_{i,j} \right\}$, where $I = \frac{w}{w_p}$, $J = \frac{h}{h_p}$ 
and $p_{i,j} \in \RR^{w_p \times h_p \times c}$ is the $(i,j)$-th patch of the
input image. The first index $i$ is the horizontal index and the other index 
$j$ is the vertical index.

First, we sweep the image vertically with two RNNs, with one RNN working in
a bottom-up direction and the other working in a top-down direction. 
Each RNN takes as an input one (flattened) patch at a time and updates its
hidden state, working \emph{along each column} $j$ of the split input image $X$.
\begin{align}
    v^F_{i,j} = f_{\text{VFWD}}(z^F_{i,j-1},p_{i,j}), &\text{ for
    }j=1,\cdots, J\\
    v^R_{i,j} = f_{\text{VREV}}(z^R_{i,j+1},p_{i,j}), &\text{ for
    }j=J,\cdots,1 
\end{align}

Note that $f_{\text{VFWD}}$ and $f_{\text{VREV}}$ return the activation of the
recurrent hidden state, and may be implemented either as a simple $\tanh$ layer,
as a gated recurrent layer~\citep{Cho2014} or as a long short-term memory
layer~\citep{Hochreiter+Schmidhuber-1997}.

After this vertical, bidirectional sweep, we concatenate the intermediate hidden 
states $v^F_{i,j}$ and $v^R_{i,j}$ at each location $(i,j)$ to get a composite 
feature map $V= \left\{ v_{i,j} \right\}_{i=1,\ldots,I}^{j=1,\ldots,J}$, where 
$v_{i,j} \in \RR^{2d}$ and $d$ is the number of recurrent units.
Each $v_{i,j}$ is now the activation of a feature detector at the location
$(i,j)$ with respect to all the patches in the $j$-th column of the original
input ($p_{i, j}$ for all $i$).

Next we sweep over the obtained feature map $V$ horizontally with two RNNs
($f_{\text{HFWD}}$ and $f_{\text{HREV}}$).
In a similar manner as the vertical sweep, these RNNs work along each row 
of $V$ resulting in the output feature map $H = \left\{ h_{i,j} \right\}$, where
$h_{i,j} \in \RR^{2d}$. Now, each vector $h_{i,j}$ represents the features of
the original image patch $p_{i,j}$ \emph{in the context of the whole image}.

Let us denote by $\phi$ the function from the input image map of $X$ to
the output feature map $H$ (see Fig.~\ref{fig:networklayer} for a graphical
illustration.) Clearly, we can stack multiple $\phi$'s to make the
proposed ReNet deeper and capture increasingly complex features of the input
image. After any number of recurrent layers are applied to an input image,
the activation at the last recurrent layer may be flattened and fed into a
differentiable classifier. In our experiments we used several 
fully-connected layers followed by a softmax classifier (as shown in Fig.~\ref{fig:network}).

The deep ReNet is a smooth, continuous function, and the parameters (those from
the RNNs as well as from the fully-connected layers) can be estimated by the
stochastic gradient descent algorithm with the gradient computed by
backpropagation algorithm~\citep[see, e.g.,][]{BP86} to maximize the
log-likelihood.

\section{Differences between LeNet and ReNet}
\label{sec:lenetrenet}

There are many similarities and differences between the proposed ReNet and a
convolutional neural network. In this section we use LeNet to refer to the
canonical convolutional neural network as shown by \citet{LeCun89}. Here we
highlight a few key points of comparison between ReNet and LeNet.

At each layer, both networks apply the same set of filters to patches of the
input image or of the feature map from the layer below. ReNet, however,
propagates information through lateral connections that span across the whole
image, while LeNet exploits local information only. The lateral connections
should help extract a more compact feature representation of the input image at
each layer, which can be accomplished by the lateral connections
removing/resolving redundant features at different locations of the image. This
should allow ReNet resolve small displacements of features across multiple
consecutive patches. 
%Also, the lack of
%this type of lateral connection in LeNet may lead to many more levels of
%convolution+pooling layers in order to detect redundant features from different
%parts of the image. 

LeNet max-pools the activations of each filter over a small region to achieve
local translation invariance. In contrast, the proposed ReNet does not use any
pooling due to the existence of learned lateral connections. The lateral
connection in ReNet can emulate the local competition among features induced by
the max-pooling in LeNet.  This does not mean that it is not possible to use
max-pooling in ReNet. The use of max-pooling in the ReNet could be helpful in
reducing the dimensionality of the feature map, resulting in lower computational
cost. 

Max-pooling as used in LeNet may prove problematic when building a
convolutional autoencoder whose decoder is an inverse\footnote{
    All the forward arrows from the input to the output in the original LeNet
    are reversed.
}
of LeNet, as the max operator is not invertible. The proposed
ReNet is end-to-end smooth and differentiable, making it more suited to be used
as a decoder in the autoencoder or any of its probabilistic variants~\citep[see,
e.g.,][]{Kingma+Welling-ICLR2014}.

In some sense, each layer of the ReNet can be considered as a variant of a usual
convolution+pooling layer, where pooling is replaced with lateral connections,
and convolution is done without any overlap. Similarly, \citet{Springenberg2014}
recently proposed a variant of a usual LeNet which does not use any pooling.
They used convolution with a larger stride to compensate for the lack of
dimensionality reduction by pooling at each layer. However, this approach still
differs from the proposed ReNet in the sense that each feature activation at a
layer is only with respect to a subset of the input image rather than the whole
input image.

The main disadvantage of ReNet is that it is not easily parallelizable, due to
the sequential nature of the recurrent neural network (RNN). LeNet, on the other
hand, is highly parallelizable due to the independence of computing activations
at each layer. The introduction of sequential, lateral connections, however, may
result in more efficient parametrization, requiring a smaller number of
parameters with overall fewer computations, although this needs to be further
explored. We note that this limitation on parallelization applies only to
model parallelism, and any technique for data parallelism may be used for both 
the proposed ReNet and the LeNet.

\section{Experiments}

\subsection{Datasets}

We evaluated the proposed ReNet on three widely-used benchmark datasets; MNIST,
CIFAR-10 and the Street View Housing Numbers (SVHN). In this section we describe 
each dataset in detail.

\paragraph{MNIST}
The MNIST dataset~\citep{Lecun99objectrecognition} consists of 70,000
handwritten digits from 0 to 9, centered on a $28\times 28$ square canvas. Each
pixel represents the grayscale in the range of $\left[0, 255\right]$.\footnote{
    We scaled each pixel by dividing it with $255$.
} 
We split the dataset into 50,000 training samples, 10,000 validation
samples and 10,000 test samples, following the standard split.

\paragraph{CIFAR-10}
The CIFAR-10 dataset~\citep{KrizhevskyHinton2009} is a curated subset of the 80
million tiny images dataset, originally released by
\citet{Torralba+Fergus+Freeman-2008}. CIFAR-10 contains 60,000 images each of which
belongs to one of ten categories; airplane, automobile, bird, cat, deer, dog,
frog, horse, ship and truck. Each image is 32 pixels wide and 32 pixels high
with 3 color channels (red, green and blue.) Following the standard procedure,
we split the dataset into 40,000 training, 10,000 validation and 10,000 test
samples. We applied zero-phase component analysis (ZCA) and normalized each pixel
to have zero-mean and unit-variance across the training
samples, as suggested by \citet{KrizhevskyHinton2009}.

\paragraph{Street View House Numbers}
The Street View House Numbers (SVHN) dataset~\citep{Netzer-wkshp-2011} consists 
of cropped images representing house numbers captured by Google StreetView 
vehicles as a part of the Google Maps mapping process. These images consist of
digits 0 through 9 with values in the range of [0, 255] in each of 3
red-green-blue color channels. Each image is 32 pixels wide and 32 pixels
high giving a sample dimensionality (32, 32, 3). The number of samples we used for
training, valid, and test sets is 543,949, 60,439, and 26,032 respectively. We
normalized each pixel to have zero-mean and unit-variance across the training
samples.

\subsubsection{Data Augmentation}

It has been known that augmenting training data often leads to better
generalization~\citep[see, e.g.,][]{Krizhevsky-2012}. We decided to employ 
two primary data augmentations in the following experiments: {\it flipping} and 
{\it shifting}. 

For flipping, we either flipped each sample horizontally with 25\% chance, 
flipped it vertically with 25\% chance, or left it unchanged. This allows lets 
the model observe ``mirror images'' of the original image during training. 
In the case of shifting, we either shifted the image by 2 pixels to the left 
(25\% chance), 2 pixels to the right (25\% chance) or left it as it was. After 
this first processing, we further either shifted it by 2 pixels to the top 
(25\% chance), 2 pixels to the bottom (25\% chance) or left it as it was. This 
two-step procedure makes the model more robust to slight shifting of an object 
in the image. The shifting was done without padding the borders of the image, 
preserving the original size but dropping the pixels which are shifted out of
the input while shifting in zeros.

The choice of whether to apply these augmentation procedures on each dataset was
chosen on a per-case basis in order to maximize validation performance. 

\begin{figure}[t]
    \advance\leftskip-0.2\columnwidth
    \centering
    \makebox[\columnwidth][c]{\includegraphics[height=.135\textheight,width=1.1\columnwidth]{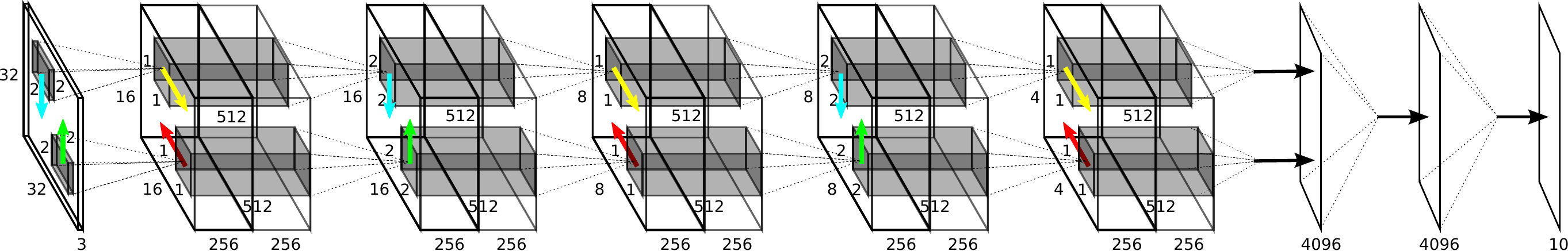}}
    \caption{The ReNet network used for SVHN classification}
    \label{fig:network}
\end{figure}

\subsection{Model Architectures}

\paragraph{Gated Recurrent Units}

Gated recurrent units~\citep[GRU,][]{Cho2014} and long short-term memory
units~\citep[LSTM,][]{Hochreiter+Schmidhuber-1997} have been successful in many
applications using recurrent neural networks~\citep[see,
e.g.,][]{Cho2014,Sutskever-et-al-NIPS2014,Xu-et-al-arxiv2015}. To show that the
ReNet model performs well independently of the
specific implementation of the recurrent units, we decided to use the GRU on 
MNIST and CIFAR-10, with LSTM units on SVHN.

The hidden state of the GRU at time $t$ is computed by
\begin{align*}
    h_t = (1-u_t) \odot h_{t-1} + u_t \odot \tilde{h}_t,
\end{align*}
where
\begin{align*}
    \tilde{h}_t = \tanh\left( W x_t + U (r_t \odot h_{t-1}) + b \right)
\end{align*}
and
\begin{align*}
    \left[ u_t; r_t\right] = \sigma\left( W_g x_t + U_g h_{t-1} + b_g\right).
\end{align*} 

For more details on the LSTM unit, as well as for an in-depth comparison among 
different recurrent units, we refer the reader to \citep{Chung2015}.

\paragraph{General Architecture}

The principal parameters that define the architecture of the proposed ReNet are 
the number of ReNet layers ($N_{\text{RE}}$), their corresponding receptive field 
sizes ($w_p \times h_p$) and feature dimensionality ($d_{\text{RE}}$), 
the number of fully-connected layers ($N_{\text{FC}}$) and their corresponding 
numbers ($d_{\text{FC}}$) and types ($f_{\text{FC}}$) of hidden units. 

In this introductory work, we did not focus on extensive hyperparameter
search to find the optimal validation set 
performance. We chose instead to focus the experiments on a small set of 
hyperparameters, with the only aim to show the potential of the proposed model.
Refer to Table~\ref{tbl:architectures} for a summary of the settings that performed 
best on the validation set of the studied datasets and to Fig.~\ref{fig:network}
for a graphical illustration of the model we selected for SVHN.

\begin{table}[t]
    \centering
    \begin{tabular}{l || c | c | c }
        & MNIST & CIFAR-10 & SVHN \\
        \hline
        \hline
    $N_{\text{RE}}$ & 2 & 3 & 3 \\
        \hline
        $w_p \times h_p$ & $[2\times 2]$--$[2 \times 2]$ & $[2\times 2]$--$[2 \times 2]$--$[2
        \times 2]$ & $[2\times 2]$--$[2 \times 2]$--$[2 \times 2]$ \\
        \hline
    $d_{\text{RE}}$ & 256--256 & 320--320--320 & 256--256--256 \\
        \hline
    $N_{\text{FC}}$ & 2 & 1 & 2 \\
        \hline
    $d_{\text{FC}}$ & 4096--4096 & 4096 & 4096--4096 \\
        \hline
    $f_{\text{FC}}$ & $\max(0, x)$ & $\max(0,x)$ & $\max(0,x)$ \\
        \hline
    Flipping & no & yes & no \\
        \hline
    Shifting & yes & yes & yes \\
    \end{tabular}
    \caption{Model architectures used in the experiments. Each row shows 
             respectively the number of ReNet layers, the size of the patches, 
             the number of neurons of each ReNet layer, the number of fully 
             connected layers, the number of neurons of the fully connected
             layers, their activation function and the data augmentation 
             procedure employed.}
    \label{tbl:architectures}
\end{table}

\subsection{Training}

To train the networks we used a recently proposed adaptive learning rate 
algorithm, called Adam~\citep{Kingma2014}. In order to reduce overfitting we 
applied dropout~\citep{Srivastava14} after each layer, including both the 
proposed ReNet layer (after the horizontal and vertical sweeps) and the 
fully-connected layers. The input was also corrupted by masking out each variable 
with probability $0.2$. Finally, each optimization run was early stopped based 
on validation error.

Note that unlike many previous works, we did not retrain the model (selected
based on the validation performance) using both the training and validation
samples. This experiment design choice is consistent with our declared goal to
show a proof of concept rather than stressing absolute performance. There are
many potential areas of exploration for future work.

\begin{table}[ht]
    \centering

    \begin{minipage}{0.45\textwidth}
        \centering

        \begin{tabular}{l |  l}
            Test Error & Model  \\
            \hline
0.28\% & \citep{DBLP:conf/icml/WanZZLF13}$\star$ \\
0.31\% & \citep{DBLP:journals/corr/Graham14}$\star$ \\ 
0.35\% & \citep{DBLP:journals/corr/abs-1003-0358} \\
0.39\% & \citep{DBLP:conf/nips/MairalKHS14}$\star$ \\
0.39\% & \citep{DBLP:journals/corr/LeeXGZT14}$\star$ \\
0.4\% & \citep{DBLP:conf/icdar/SimardSP03}$\star$ \\
0.44\% & \citep{DBLP:journals/corr/Graham14a}$\star$ \\
0.45\% & \citep{Goodfellow2013}$\star$ \\
\bf{0.45\%} & \bf{ReNet} \\ 
0.47\% & \citep{Lin2014}$\star$ \\
0.52\% & \citep{DBLP:journals/pami/AzzopardiA13} \\
        \end{tabular}

        \vspace{2mm}
        (a) MNIST
    \end{minipage}
    \hfill
    \begin{minipage}{0.45\textwidth}
        \centering 

        \begin{tabular}{l |  l}
            Test Error & Model  \\
            \hline
4.5\% & \citep{DBLP:journals/corr/Graham14a}$\star$ \\
6.28\% & \citep{DBLP:journals/corr/Graham14}$\star$ \\ 
8.8\% & \citep{Lin2014}$\star$ \\
9.35\% & \citep{Goodfellow2013}$\star$ \\
9.39\% & \citep{DBLP:journals/corr/SpringenbergR13}$\star$ \\
9.5\% & \citep{DBLP:conf/nips/SnoekLA12}$\star$ \\
11\% & \citep{Krizhevsky-2012}$\star$ \\
11.10\% & \citep{DBLP:conf/icml/WanZZLF13}$\star$ \\
\bf{12.35\%} & \bf{ReNet} \\
15.13\% & \citep{DBLP:journals/corr/abs-1301-3557}$\star$ \\
15.6\% & \citep{DBLP:journals/corr/abs-1207-0580}$\star$ \\
        \end{tabular}

        \vspace{2mm}
        (b) CIFAR-10
    \end{minipage}

    \vspace{4mm}
    \begin{minipage}{0.45\textwidth}
        \centering
        \begin{tabular}{l |  l}
            Test Error & Model  \\
            \hline
1.92\% & \citep{DBLP:journals/corr/LeeXGZT14}$\star$ \\
2.23\% & \citep{DBLP:conf/icml/WanZZLF13}$\star$ \\
2.35\% & \citep{Lin2014}$\star$ \\
\bf{2.38\%} & \bf{ReNet} \\
2.47\% & \citep{Goodfellow2013}$\star$ \\
2.8\% & \citep{DBLP:journals/corr/abs-1301-3557}$\star$ \\
        \end{tabular}

        \vspace{2mm}
        (c) SVHN
    \end{minipage}
    \hfill
    \begin{minipage}{0.45\textwidth}
        \caption{Generalization errors obtained by the
            proposed ReNet along with those reported by previous works
            on each of the three datasets. $\star$ denotes a
            convolutional neural network. We only list the results reported by a single
            model, i.e., no ensembling of multiple models.
            In the case of SVHN, we report results from models trained on 
            the Format 2 (cropped digit) dataset only.}
        \label{tbl:result}
    \end{minipage}
\end{table}

\section{Results and Analysis}

In Table~\ref{tbl:result}, we present the results on three datasets, 
along with previously reported results. 

It is clear that the proposed ReNet performs comparably to deep convolutional
neural networks which are the {\it de facto} standard for object recognition.
This suggests that the proposed ReNet is a viable alternative to convolutional
neural networks (CNN), even on tasks where CNNs have historically dominated.
However, it is important to notice that the proposed ReNet does not outperform
state-of-the-art convolutional neural networks on any of the three benchmark
datasets, which calls for more research in the future. 

\section{Discussion}

\paragraph{Choice of Recurrent Units}
Note that the proposed architecture is independent of the chosen recurrent
units. We observed in preliminary experiments that gated
recurrent units, either the GRU or the LSTM, outperform a usual sigmoidal unit
(affine transformation followed by an element-wise sigmoid function.) This
indirectly confirms that the model utilizes long-term dependencies across an
input image, and the gated recurrent units help capture these dependencies.

\paragraph{Analysis of the Trained ReNet}
In this paper, we evaluated the proposed ReNet only quantitatively. However, the
accuracies on the test sets do not reveal what kind of image structures the
ReNet has captured in order to perform object recognition. Due to the large
differences between ReNet and LeNet discussed in
Sec.~\ref{sec:lenetrenet}, we expect that the internal behavior of ReNet
will differ from that of LeNet significantly. Further investigation along
the line of \citep{ZeilerFergus14} will be needed, as well exploring ensembles
which combine RNNs and CNNs for bagged prediction.

\paragraph{Computationally Efficient Implementation}
As discussed in Sec.~\ref{sec:lenetrenet}, the proposed ReNet is less
parallelizable due to the sequential nature of the recurrent neural network
(RNN). Although this sequential nature cannot be addressed directly, our
construction of ReNet allows the forward and backward RNNs to be run
independently from each other, which allows for parallel computation.
Furthermore, we can use many parallelization tricks widely used for training
convolutional neural networks such as parallelizing fully-connected layers
~\citep{krizhevsky2014one}, having separate sets of kernels/features in
different processors~\citep{Krizhevsky-2012} and exploiting data parallelism.  

\section*{Acknowledgments}

The authors would like to thank the developers of
Theano~\citep{Bergstra2010,Bastien2012}. We acknowledge the support of the
following organizations for research funding and computing support: NSERC,
Samsung, Calcul Qu\'{e}bec, Compute Canada, the Canada Research Chairs and
CIFAR.  F.V. was funded by the AI*IA Young Researchers Mobility Grant
and the Politecnico di Milano PHD School International Mobility Grant.

\small
%\bibliography{myrefs,ml}
\bibliography{myrefs}

\end{document}